\documentclass[9pt,twocolumn,twoside]{pnas-for-arXiv}

\templatetype{pnasresearcharticle} 

\title{Quantifying the Roles of Visual, Linguistic, and Visual-Linguistic Complexity in Verb Acquisition}

\author[a,1]{Yuchen Zhou}
\author[a,b,c]{Michael J. Tarr}
\author[a]{Daniel Yurovsky} 

\affil[a]{Department of Psychology, Carnegie Mellon University, Pittsburgh, PA 15213}
\affil[b]{Neuroscience Institute, Carnegie Mellon University, Pittsburgh, PA 15213}
\affil[c]{Machine Learning Department, Carnegie Mellon University, Pittsburgh, PA 15213}

\leadauthor{Zhou}

\significancestatement{
Numerous studies have explored why verb meanings are harder to learn than nouns. Previous investigations have examined whether this challenge stems from the structure of categories in the world, the structure of language itself, or interactions between the two. We propose a novel approach that integrates these perspectives and offers a way to quantitatively measure the relative impact of each factor on early word acquisition. We find that all three sources contribute, but the primary obstacle of verb meaning acquisition is visual variability. Our framework not only explains the challenge of verb learning, but also may help account for other phenomena in language acquisition and category learning.}

\authorcontributions{Y.Z. and D.Y. designed research; Y.Z. performed research; Y.Z. and D.Y. analyzed data; Y.Z., M.J.T., and D.Y. wrote the paper.}
\authordeclaration{The authors declare no competing interest.}
\correspondingauthor{\textsuperscript{1}To whom correspondence should be addressed. E-mail: zhouyuchen@cmu.edu}


\begin{abstract}
Children typically learn the meanings of nouns earlier than the meanings of verbs. However, it is unclear whether this asymmetry is a result of complexity in the visual structure of categories in the world to which language refers, the structure of language itself, or the interplay between the two sources of information. We quantitatively test these three hypotheses regarding early verb learning by employing visual and linguistic representations of words sourced from large-scale pre-trained artificial neural networks. Examining the structure of both visual and linguistic embedding spaces, we find, first, that the representation of verbs is generally more variable and less discriminable within domain than the representation of nouns. Second, we find that if only one learning instance per category is available, visual and linguistic representations are less well aligned in the verb system than in the noun system. However, in parallel with the course of human language development, if multiple learning instances per category are available, visual and linguistic representations become almost as well aligned in the verb system as in the noun system. Third, we compare the relative contributions of factors that may predict learning difficulty for individual words. A regression analysis reveals that visual variability is the strongest factor that internally drives verb learning, followed by visual-linguistic alignment and linguistic variability. Based on these results, we conclude that verb acquisition is influenced by all three sources of complexity, but that the variability of visual structure poses the most significant challenge for verb learning.
\end{abstract}

\dates{This manuscript was compiled on \today}

\begin{document}

\maketitle
\thispagestyle{firststyle}
\ifthenelse{\boolean{shortarticle}}{\ifthenelse{\boolean{singlecolumn}}{\abscontentformatted}{\abscontent}}{}

Children's early vocabularies are dominated by nouns \cite{bates1994developmental}. 
Although there is some variability in the size of the effect across cultures and languages, children learn nouns more quickly than verbs and other predicates \cite{fenson1994variability, frank2021variability}. Attempts to understand the challenges of verb learning have generally focused on determining whether the primary source of difficulty is the structure of categories in the world to which the language refers, the structure of the language in and of itself, or the interplay between visual and linguistic information.

To know the meaning of a word, learners need to identify the consistencies among objects or actions in the world that are labeled by the word, that is, to learn the perceptual category to which each word refers \cite{peirce1974collected, quine1960word}. Consequently, much of the work in early word learning concerns the cognitive processes children use to map words onto visual categories \cite{landau1988importance, markman1988children, waxman1995words}. As such, one class of explanations for the challenges of verb learning focuses on the relative difficulty of visual category learning: referents labeled by verbs may be more variable or more confusable than those labeled by nouns \cite{gentner1982nouns, golinkoff2008toddlers}. 
For example, ``dogs'' differ from each other on a variety of dimensions such as color, size, and fur texture, but share a common basic shape \cite{Rosch1976}. In contrast, ``walks'' differ from each other not only in the identity of the agent doing the walking but also in dimensions such as speed, direction, number of limbs, and gait.

Linguistic contexts formed by other words in which a target word appears are also an informative source for learning the meaning of the target word. Deriving from Firth's intuition that ``you shall know a word by the company it keeps'' \cite{firth1957synopsis}, congenitally blind children can correctly name the colors of common objects and even the similarities among these colors, even though color is a purely visual property \cite{landau2009language}. Consequently, another class of explanations for the challenges of verb learning has focused on the linguistic contexts of words. Such explanations attribute the difficulty of verb learning to the fact that verbs refer to relations among nouns and thus are likely to occur in more diverse linguistic contexts. Supporting this account, children learn verbs that occur in more restricted contexts earlier than verbs that occur in more diverse contexts \cite{goldberg2004learning}.

Finally, these visual and linguistic explanations are not mutually exclusive: the difficulty of learning verbs may emerge from interactions between these two sources of information. That is, verbs are more difficult to learn than nouns because their linguistic usage is less transparently related to the events with which they occur. For example, English tends to encode the manner of the motion into the verb (e.g., as in ``The bottle floated into the cave.''), while Spanish tends to lexicalize the direction of the movement when describing the same event, leaving manner as an optional adjunct (e.g., as in ``La botella entró en la cueva, flotando.'', meaning ``The bottle entered the cave floating.'') \cite{gentner1982nouns}. Furthermore, the contexts of transitive verbs are fundamentally ambiguous unless the words for their agent and patient are known (e.g., as in ``chase'' and ``flee'') \cite{gleitman1990structural}. Accordingly, verbs may require more complex reasoning about making a selection of the available relationships and identifying the syntactic structure of the utterances.

To date, efforts to compare these explanations have relied on indirect methods such as cross-linguistic comparisons \cite{talmy1975semantics}, studies of atypical learners \cite{snedeker2007starting, goldin1977development}, or artificial language learning experiments that try to approximate the real-world learning problems faced by children \cite{gillette1999human}. 
Here we introduce a unified method for quantifying the complexity of both the visual and linguistic contexts in which a word appears, as well as the coherence of both sources of information. By independently measuring these three sources of information for word learning, we can directly evaluate them as explanations for the challenge posed by verb learning. 

\begin{figure*}
\centering
\includegraphics[width=\textwidth]{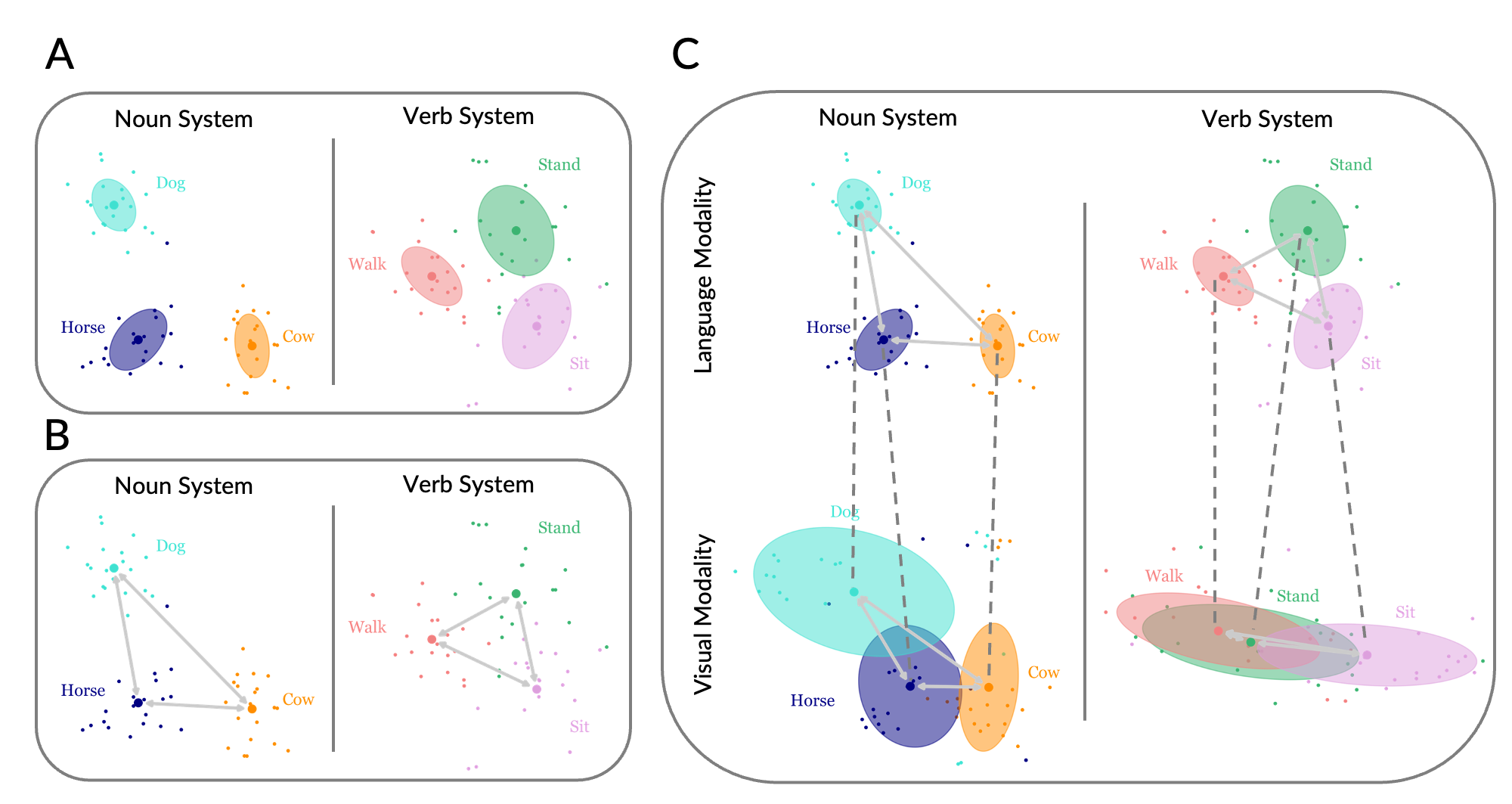}
\caption{Potential sources of difficulty for learning nouns and verbs as illustrated by selected concepts in embedding spaces (see Methods for visualization details). Large dots represent category centroids, small dots represent learning instances of corresponding categories, and colored ellipses represent 68\% confidence (one standard deviation) areas for each category. (A) Verb exemplars within each category are more variable as compared to noun exemplars within each category. (B) Verb categories overlap more and are less distinguishable from one another as compared to noun categories. (C) For verbs, the structures of linguistic and visual representations are inconsistent across modalities. For nouns, the structures of linguistic representations and visual representations are better aligned.}
\label{fig:schematic}
\end{figure*}

\section*{Measuring Visual and Linguistic Complexity}

To understand how visual information, linguistic information, and their interactions contribute to differences in the learnability of nouns and verbs, it is critical to represent them in a unified manner.
Historically, efforts to understand the categorical structure of meanings in psychology have relied on top-town approach, that is, defining the sets of features on which meanings can vary. For example, in studies of category learning, these features are often explicitly defined and manipulated by experimenters and include properties such as color, size, orientation, and shape \cite{shepard1961learning, ashby2005human}. Another example can be found in studies of language learning, where these features are often generated by human annotators based on their intuitions about the core properties of the meaning of a word \cite{mcrae2005semantic, ameel2005bilinguals, colunga2005lexicon}.

In contrast, because of the arbitrary relationship between the form of a word and its meaning \cite{de1989cours}, a bottom-up approach is taken  in studies of the structure of individual word meanings, where word meanings emerge from the statistical relationships between words themselves. 
In particular, models that define a word by its frequency of co-occurrence with other words rose in prominence in the 1990s because of their ability to account for human similarity judgments and to solve simple analogies \cite{lund1996producing, landauer1997solution}. 

Modern language models build on similar intuitions regarding word co-occurrences, but with richer representations, more powerful learning algorithms, and far larger and more diverse datsets. These larger models such BERT and even more powerful, recent ``large language models'' (LLMs) have exploded into prominence for both research and real-world applications \cite{mikolov2013efficient, pennington2014glove, devlin2018bert, brown2020language,ouyang2022training}. 
The representations learned by modern language models have been shown to generate similarity judgments aligned with human similarity judgments \cite{hill2015simlex}, to predict human behavior in semantic fluency tasks \cite{hills2012optimal}, to predict choices in a variety of classical decision-making tasks \cite{bhatia2017associative}, and to encode many of the top-down features identified by linguists as characterizing the properties of natural language. As such, these models appear to be effective at capturing the meanings of words.

We leverage these recent advances in large-scale pre-trained language models to understand the structure of meanings that words represent. In addition to generating word embeddings by the language model, we similarly use a pre-trained vision model that was trained in an unsupervised fashion to generate visual embeddings from pixels that represent the visual structure of the categories to which words refer. Because both the linguistic and visual dimensions of words' meanings are encoded in the same kind of representation, we can then apply the same measures to examine whether the later onset of verb learning arises from one or more of the following factors: 1) identifying stable visual categories, 2) understanding meanings from linguistic contexts, and 3) aligning the visual and linguistic structures corresponding to the same word. 

Intuitively, the meaning of a word might be hard to learn from the contexts in which it appears in for two different reasons. First, different instances of a word may be very different from one another.
Such variability makes it difficult to converge on the central meaning of a word. Second, two different words may appear in very similar contexts, making the two words difficult to distinguish from one another. A natural method for quantifying these two sources of difficulty is to consider both the similarities between multiple instances of the same word and the dissimilarities between different words. As illustrated in Figure \ref{fig:schematic}A, the distribution of verbs may be more scattered compared to that of nouns. At the same time, as illustrated in Figure \ref{fig:schematic}B, verb categories may be generally less dissimilar to one another than noun categories.

Given common representations, we are also able to measure the alignment between the linguistic representations of words and the visual representations of their corresponding categories (Figure \ref{fig:schematic}C). Prior work has demonstrated that representations across these two modalities tend to be highly aligned for concrete nouns, and further, that the earliest learned words are more highly-aligned than later learned words \cite{roads2020learning}. We explore whether the degree of alignment between linguistic and visual meanings of words helps distinguish noun learning from verb learning. For nouns, we hypothesize that the visual structure will be more similar to the linguistic structure, while for verbs we hypothesize that the visual structure will be less similar to the linguistic structure.

Finally, though these measures may all be able to explain the difficulty of word learning to a greater or lesser extent, we are interested in what are the main obstacles that children need to overcome to successfully acquire the meanings of words. Accordingly, we estimate the relative contribution of each of the measures to word learning in a single regression analysis by asking how well each factor can predict Age of Acquisition (the time at which a word is learned).

\section*{Results}


As the first stage of the analysis, we examine the structure of categories in each modality by using unsupervised machine learning representations of 210 nouns and 210 verbs from the Visual Genome dataset (see Methods for details) to probe how easy it is for a simulated language learner to form new categories and tell apart two categories. As such, the structure of categories can be captured by two measures: how dispersed each individual category is, and how far categories are from one another. Accordingly, we quantify the structure of representational space by two metrics: the variability of individual categories and the discriminability of different categories (see Methods for mathematical definitions).

\begin{figure}
\centering
\includegraphics[width=\linewidth]{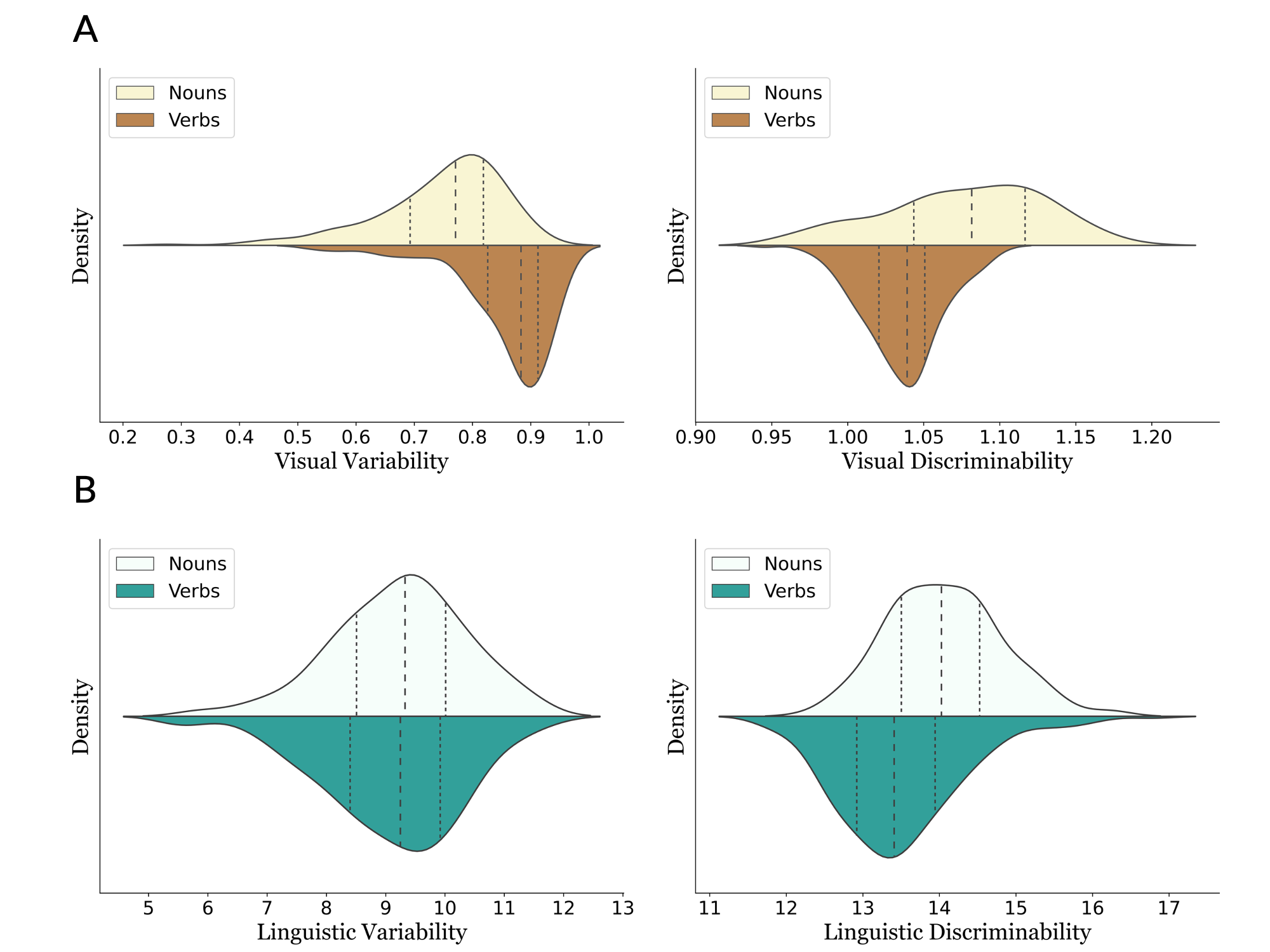}
\caption{(A) Distributions of variability and discriminability of nouns and verbs from the Visual Genome dataset in the visual modality. Verb categories are more variable and less distinguishable than noun categories. (B) Distributions of variability and discriminability of nouns and verbs from the Visual Genome dataset in the language modality. Verb categories are equally variable but less distinguishable than noun categories. Dashed lines denote 25th-, 50th-, and 75th-percentile of distributions.}
\label{fig:var_dis}
\end{figure}

\subsection*{Category Structure in Visual Modality}
As shown in Figure \ref{fig:var_dis}A, the visual variability of verbs is significantly greater than that of nouns, $t(418)=11.71$, $p<0.001$, and the visual discriminability of verbs is significantly lower than that of nouns, $t(418)=-9.82$, $p<0.001$. Thus, consistent with our hypothesis, objects labeled by the same noun are more like each other than events labeled by the same verb, and events described by different verbs are more confusable with each other than objects described by different nouns.

\subsection*{Category Structure in Language Modality}
 Although the difference in linguistic variability is not statistically significant between nouns and verbs, $t(418)=-1.15$, $p=0.250$, the linguistic discriminability of verbs is reliably lower than the linguistic discriminability of nouns, $t(418)=-7.63$, $p<0.001$ (Figure \ref{fig:var_dis}B). In other words, the linguistic meanings of verbs are not more diverse but less distinguishable compared to that of nouns. 

In sum, the observations from unimodal analyses provide us with a potential explanation of why verbs are harder to learn than nouns, i.e., compared to nouns, it is harder to capture the central meanings of verbs' visual and linguistic categories and to distinguish two verbs' visual categories. 

\begin{figure*}[ht]
\centering
\includegraphics[width=\textwidth]{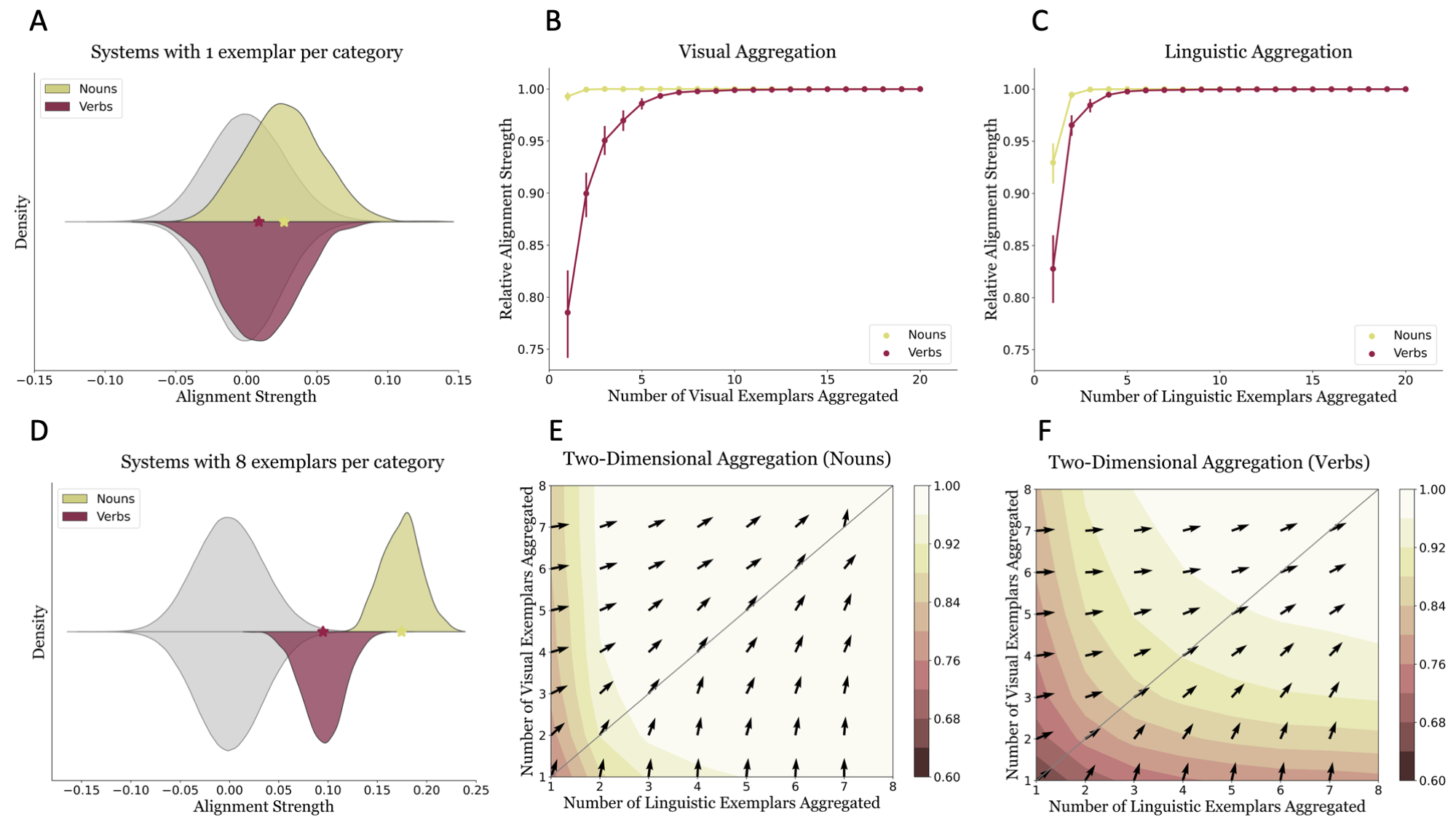}
\caption{(A) Distributions of alignment strengths of true mappings (in colors) and randomly permuted mappings (in grey). Asterisks represent the mean alignment strengths of the true mappings. Nouns are more alignable than verbs in terms of both absolute alignment strength and relative alignment strength (see main text). As visual exemplars (B) or linguistic exemplars (C) aggregate, relative alignment strength, which measures the percentage of permuted mappings that are less well aligned than the true mapping, increases and gradually converges to 1.0. The noun system is more alignable than the verb system in early aggregation stages, but the verb system becomes almost as alignable as the noun system with a sufficient number of learning instances. Error bars represent 95\% confidence intervals computed by bootstrapping over 1,000 simulations at each level. (D) Alignment strengths of true mappings (asterisks) and the distribution of permuted mappings (in grey) in systems with 8 linguistic exemplars and 8 visual exemplars aggregated per category. The alignment strength improves considerably as compared to that in the single-exemplar condition (as in A) for both nouns and verbs. (E, F) Relative alignment strength as a function of both the number of linguistic exemplars and the number of visual exemplars. Nouns can become highly alignable with only 1 visual exemplar and sufficient linguistic exemplars per category. In contrast, sufficient visual exemplars and sufficient linguistic exemplars are both necessary for the formation of a well-aligned verb system. Directions of arrows indicate directions of gradients, which represent the optimal combination of visual exemplars and linguistic exemplars that increases alignment strength most efficiently at each location.}
\label{fig:alignment}
\end{figure*}

\subsection*{Alignment Between Visual and Language Modalities}

Another potential difficulty in learning verbs, as we discussed above, is that verbs make a selection of the available relational information such that the learner has to identify which aspects of information from physical events are encoded in the verb categories. Therefore, we examine the coherence between the visual and language modalities of nouns and verbs, that is, whether visual categories can be readily mapped onto linguistic categories. We term this metric Alignment Strength (see Methods for mathematical definition). To make meaningful comparisons across systems with different words, we also measure the relative strength of alignment, that is, how well-aligned a mapping is compared to alternative permuted mappings in which linguistic and visual representations of different words are randomly matched.

Figure \ref{fig:alignment}A shows the distribution of alignment strengths of true mappings and permuted mappings in systems with one visual exemplar representing each visual category and one linguistic exemplar representing each linguistic category. 
The mean alignment strength of true mappings significantly deviates from the mean of permuted mappings for both nouns, $t(1,000,998)=30.35, p<0.001$, and verbs, $t(1,000,998)=11.17,p<0.001$, suggesting that true mappings are reliably more alignable than randomly permuted mappings.

With respect to our hypothesis, the mean alignment strength of true mappings for nouns ($\rho=0.027$) is higher than $75.6\%$ of permuted mappings, while the mean alignment strength of true mappings for verbs ($\rho=0.009$) is higher than $59.8\%$ of permuted mappings. Therefore, consistent with our hypothesis, it is more difficult to establish correct mappings from verb visual categories to corresponding words than that of nouns measured by both absolute alignment strength and relative alignment strength. These results capture the critical difference in noun and verb learning, 
indicating that identifying the correct mapping between visual categories and linguistic categories may be another vital source of difficulty to overcome in verb learning.

\subsection*{Improving Alignment by Exemplar Aggregation}

One explanation for the lower alignment strength observed for verbs is that the similarities of visual events labeled by verbs are inherently different from the similarities of their linguistic usage. However, the challenge in aligning the visual and linguistic representations of verbs may also arise from the high variability and low discriminability of verb categories such that the meaning of a verb may not be well represented by a single learning instance. In line with how humans gradually accumulate learning instances to acquire words, we aggregate exemplars by taking their arithmetic mean to examine the change of alignment strength as a function of the number of exemplars.

For visual aggregation, we form well-learned stable linguistic prototypes by aggregating 20 linguistic exemplars per category and then calculate the relative alignment strength of the system as a function of the number of visual exemplars (Figure \ref{fig:alignment}B). Because the true mapping of nouns already has higher alignment strength than almost all of the permuted mappings even with just one visual exemplar per category, averaging multiple exemplars together produced little improvement in its relative strength. In contrast, although alignment strength is initially quite low for verbs, the relative alignment strength of the verb system progressively increases as more visual exemplars aggregate and approaches the relative strength of the noun system at the point at which 8 visual exemplars per category are aggregated. 

Mirroring visual aggregation, the alignment strengths of both nouns and verbs improve as linguistic exemplars are aggregated (Figure \ref{fig:alignment}C). Different from visual aggregation, however, we find that in this setting one linguistic exemplar per category no longer produces a good mapping for the noun system. In fact, three or more linguistic exemplars per category are needed to build a well-aligned noun mapping. This difference in the number of required exemplars between visual aggregation and linguistic aggregation is consistent with empirical observations from previous studies: 3- and 4-month-old infants can discriminate two basic-level visual categories (i.e., cats and horses) after 3 exposures to each stimulus image \cite{eimas1994studies}, while 3-and 4-year-old children, which are presumably more capable, cannot fully understand the meanings of novel words after 12 exposures to the words in storybooks \cite{horst2011get}.
For verbs, similar to our findings from visual aggregation, aggregating linguistic exemplars drives the relative alignment strength from approximately 0.7 to near perfect alignment.


Absolute alignment strength is also improved via exemplar aggregation. The mean alignment strength of 1,000 randomly sampled noun systems with 8 visual and 8 linguistic exemplars per category ($\rho=0.175$)  (Figure \ref{fig:alignment}D) is significantly greater than the mean of 1,000 random noun systems with 1 visual and 1 linguistic exemplar per category ($\rho=0.027$) (Figure \ref{fig:alignment}A), $t(1998)=158.93,p<0.001$. Similarly, the mean  alignment strength of 1,000 randomly sampled verb systems with 8 visual and 8 linguistic exemplars per category ($\rho=0.095$) is significantly greater than the mean of 1,000 random verb systems with 1 visual and 1 linguistic exemplar per category ($\rho=0.009$), $t(1998)=89.74,p<0.001$.


Two-dimensional aggregation simulations in which both visual and language exemplars are aggregated provide us with a global view of the learning process (Figure \ref{fig:alignment}E, \ref{fig:alignment}F). In addition to confirming observed findings that noun systems are more alignable than verb systems in early stages and that verb systems become well-aligned with sufficient learning instances, we also find different patterns in the importance of introducing new visual learning instances versus linguistic instances: verb systems demonstrate a quasi-symmetric pattern (e.g., the alignment strength of the verb system with 1 visual exemplar + 8 linguistic exemplars is close to that of 8 visual exemplars + 1 linguistic exemplar), whereas the noun system displays a strong bias towards linguistic input (e.g., the alignment strength of the noun system with 1 visual exemplar + 8 linguistic exemplars is way better than that of the noun system with 8 visual exemplars + 1 linguistic exemplar). This observation is consistent with the natural partitions hypothesis \cite{gentner1982nouns, gentner1981some},  which posits that nouns are easier to individualize in the world and therefore the difficulty of noun learning relies less on learning corresponding visual categories.

In conclusion, aggregating multiple exemplars robustly improves the alignment strengths for both nouns and verbs. Importantly, consistent with the success of children learners over time, with sufficient learning exemplars, verbs' visual and linguistic meanings can be transparently mapped. Strikingly, this result suggests that, opposite to what we hypothesized, the difficulty in verb learning largely lies in the challenges of extracting invariant categorical representations of verbs rather than any inherent global misalignment between visual events and word usage.\footnote{Replication on a second image dataset yields similar qualitative results. We also replicate our analyses using a video dataset to better account for the dynamics of verbs. Not surprisingly, we find that learning from videos is more efficient than learning from images. See Supporting Information for details.}

\subsection*{Unpacking the Contribution of Factors}

Our analyses demonstrate that multiple features, including variability, discriminability, and alignment strength, appear to reflect important characteristics of early word learning. Additionally, a wide variety of earlier studies have documented that word frequency is a significant predictor of when a word will be learned \cite{manybabies2020quantifying, goodman2008does, ebbinghaus2013memory, brown1987first, ambridge2015ubiquity}. To answer the ultimate question of what drives early word learning, we use a type of gradient-boosted trees model called XGBoost \cite{chen2016xgboost} to run a regression analysis that takes into account all of the factors that might be associated with word learning including word frequency and word type to predict the learnability of individual words. Age of Acquisition (AoA), which is the age at which a word is typically learned, is measured on child-directed speech corpus as a proxy of the learnability of words. 

\begin{figure*}
    \centering
\includegraphics[width=\linewidth]{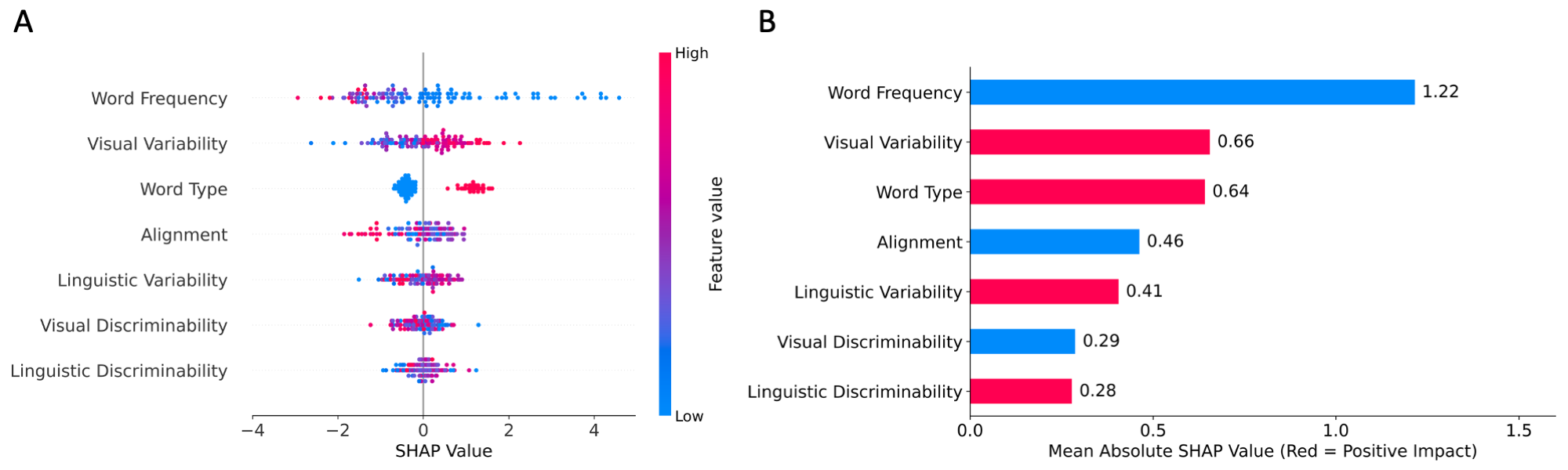}
\caption{An XGBoost model was trained to predict the learnability of words as measured by Age of Acquisition (AoA). SHAP values were computed to interpret the relative contribution of each feature. (A) Summary of the importance of all features in predicting AoA. Each dot represents one word category. If the SHAP value of a category with regard to a particular feature is positive, then this feature positively impacts the predicted variable in this observation, and vice versa. For instance, frequency negatively influences AoA because high word frequency generally corresponds to a low SHAP value. Note that ``type'' refers to word type with verbs encoded as 1 and nouns encoded as 0. (B) Global feature importance quantified by the mean absolute SHAP values of each feature across all word categories. Features that are positively correlated with the predicted variable (AoA) are colored in red, while features that are negatively correlated with AoA are colored in blue. Frequency, visual variability, word type, alignment strength, and linguistic variability are all good predictors of AoA.}
\label{fig:shap}
\end{figure*}

For better interpretability of results, we adopt SHAP \cite[SHapley Additive exPlanations,][]{lundberg2017unified} method to visualize the predictions made by the model. Figure \ref{fig:shap}A illustrates the contribution of each factor in predicting AoA in a bee swarm plot. Each dot corresponds to one word category, and the position of the dot regarding 0 on the x-axis reflects whether this factor positively or negatively contributes to the value of AoA. Consistent with prior studies, we find that frequency is a reliable predictor of AoA: words with high frequencies typically have low AoAs. Word type is also distinctly associated with the learnability of words: without other information, knowing a word is a verb always increases the expected AoA of the word.

Because the contribution of other factors is less obvious, we compute the global importance of each factor by calculating the mean of absolute SHAP values of each factor across all words (Figure \ref{fig:shap}B). All factors but linguistic discriminability are correlated with AoA in the way that we hypothesized: word frequency, alignment, and visual discriminability positively impact AoA, while visual variability, word type, and linguistic variability negatively impact AoA. Interestingly, visual variability is such a strong predictor of AoA that it is comparable to word type in magnitude. In contrast, both visual and linguistic discriminability show little contribution in predicting AoA, which may account for the incorrect prediction of the model with respect to the impact of linguistic discriminability.

Comparing the effects of factors with benchmarks (word frequency and word type), we conclude that visual variability is the most important feature among factors we considered that internally drives early word learning. Alignment and linguistic variability are also reliably associated with the learning difficulty of words, while neither visual discriminability nor linguistic discriminability shows a distinct effect on predicting when a word will be learned.


\section*{General Discussion}




A long-standing question in the study of language development is why verbs are harder to learn than nouns. 
In this paper, we systematically examine the relative contribution of three sources of information: (1) the structure of visual categories in the world to which language refers, (2) the structure of language itself, and (3) the interplay between visual and linguistic information. Supporting the natural partitions hypothesis, which states that nouns associated with concrete objects are easy to acquire because they are naturally consistent and individuated referents \cite{gentner1981some, gentner1982nouns, gentner2001individuation}, we demonstrate that: (1)~robust mappings between nouns and corresponding visual objects can be established with just one encounter with the visual object; (2)~robust mappings between nouns and corresponding visual objects can be established with a relatively small number of encounters with linguistic instances for each noun; (3)~language processing, rather than visual object identification, is the main obstacle to overcome for noun learning. In contrast, verbs associated with dynamic events are held to be harder to acquire because they are more variable. That is, the actions themselves and the identity of the agents performing the actions and the patients receiving the actions vary across different contexts. Supporting this hypothesis, an examination of embedding spaces demonstrates that verbs are more visually variable than nouns. More generally, we find that all three sources of information contribute to the difficulty of word learning, but visual variability is the most predictive internal factor of the age at which children learn individual words -- a finding that conforms to earlier observations that more imaginable concepts (evoking mental images more easily) are acquired more rapidly and earlier in development \cite{mcdonough2011image, gentner2001individuation}.

Another prevailing theoretical claim about verbs states that verbs are hard to learn because they make a selection of both particular objects in the environment and the relationships between those objects. This is different from nouns, which involve a direct mapping between words and visual objects. Reflecting this difference between verbs and nouns, there is evidence that the selection of particular properties of actions emphasized by verbs varies across languages while nouns are relatively consistent cross-linguistically~\cite{gentner1981some, gentner1982nouns, talmy1975semantics}. One implication of this theory is that children need to put more ``effort'' into identifying the core properties of verb categories as learn a language (e.g., attending more to the structure of visual examples of verbs and to the larger context of linguistic examples). However, our results suggest that increased attention or other forms of effort to identify core properties of verbs are not necessarily a difficulty to the effective acquisition of verb meanings. We demonstrate that mapping referents with verbs can be resolved by exemplar aggregation, suggesting that the main difficulty in learning verbs lies in the challenge of extracting invariant categorical representations of verbs rather than any inherent global misalignment between visual events and verb usage. In other words, the way by which the English linguistic space is carved up is not arbitrary and is in fact similar to how events visually appear to be. To better understand what pieces of relational information are selected to lexicalize events and how that may affect language learning, cross-linguistic comparison is in need. It is also possible that alignment strength as a global measurement is not sensitive to local distortions (children may confuse ``push'' with ``pull'' but are less likely to confuse ``push'' with ``run''), so more fine-grained analyses would help answer the question.


Unlike previous studies on early language acquisition that generally relied on a limited number of case studies or in-lab experiments where stimuli were designed and manipulated manually, we sample learning instances of a relatively large number of words from large corpora and use artificial neural networks that were trained in an unsupervised fashion to allow features to emerge bottom-up. Most importantly, by encoding visual and linguistic information in the same kind of representation, our quantitative analyses provide an integrative view, rather than an all-or-nothing mindset, on the difficulties that children face when acquiring nouns and verbs, which may open a new door for resolving the long lasting debates of why verbs are harder to learn than nouns.

Undoubtedly, the mechanisms underlying children's rapid word learning are more complex than what we have considered in our analyses. This paper has a main focus on word-level semantic learning and does not take into account other documented factors including inflectional morphology \cite{brown2013first}, syntactic constructions \cite{ninio1999pathbreaking}, and how easy it is to segment a word from a continuous speech stream \cite{saffran1996statistical}. Moreover, children are regarded as passive learners in our analysis as we only examine language comprehension, but factors such as interactions between children and parents/caregivers \cite{nelson2007young, adamson2018communication} and children's intentions \cite{meltzoff1995understanding, gergely1995taking} also play their roles in guiding language learning. However, we believe this quantitative framework is adjustable and can be applied to model a wide range phenomena in language development and broad concept learning.

In conclusion, we answer the question of why verbs are harder to learn than nouns, or more generally, why some words are harder to learn than others, by systematically analyzing the visual, linguistic, and visual-linguistic alignment challenges children face during early language acquisition. Our results robustly demonstrate that the main difficulty in word learning is to properly identify visual objects or events that can be described by the same word. Moreover, identifying consistent linguistic contexts in which the same word appears and establishing mappings between visual events and word usage are the other two puzzles that children need to solve in order to successfully acquire language.



\matmethods{\subsection*{Datasets} Visual Genome (VG) was used as the dataset to sample visual images of nouns and verbs \cite{krishnavisualgenome}. Images in VG have dense annotations of objects and relationships between objects in the form of Wordnet Synset \cite{miller1995wordnet}, which lemmatizes synonymous concepts. We selected the 210 verbs that appear no less than 20 times in the whole dataset to form the verb system. Because nouns are generally more frequent than verbs, we selected the most frequent 210 nouns to ensure the size of the noun system is the same as the size of the verb system. The region of interest (ROI) of a noun is determined by the bounding box of the annotated object, and the ROI of a verb is determined by the union of the bounding boxes of any agents and/or objects that take part in the verb.

Visual representations were derived from a neural network based on ResNet-50 architecture \cite{he2016deep}. The network was pre-trained in an unsupervised fashion by an algorithm called Swapping Assignments between Views on the ImageNet ILSVRC2012 dataset \cite{caron2020unsupervised}. The unsupervised training paradigm ensures that the model was trained only to encode low-level visual features (e.g., colors, dots, edges) and is not biased towards specific categories in the pretraining dataset so that the model can serve as a proxy of the human visual system.

For language data, we sampled texts that contain the 210 nouns/verbs from Wikipedia articles. We defined the context of a word to be the 25 words that precede the target word and the 25 that follow it, which sums up to a window of 51 words. Post-hoc experiments showed that results are not sensitive to window sizes ranging from 5 to 101.

Bidirectional Encoder Representations from Transformers \cite[BERT,][]{devlin2018bert} was employed to extract language representations of words because of its ability to produce contextual-sensitive embeddings as the contexts of the target word change. We obtained the pretrained uncased BERT model from the Transformers library \cite{wolf2019huggingface}.

\subsection*{Visualization in Schematic Diagram}
We used representations from real data to illustrate our hypotheses. For each category, 20 exemplars were sampled and t-SNE was applied to project representations to a two-dimensional plane. Data were assumed to follow 2D Gaussian distributions for the estimation of parameters of ellipses. Embeddings in the same modality were plotted on the same scale for visual comparison. Linguistic representations from BERT were taken as an example for demonstration in panel A and panel B.

\subsection*{Variability and Discriminability}
We quantified the structure of embedding spaces by two metrics: the variability of individual categories, which is computed as the average Euclidean distance from exemplars to category centroids, and the discriminability of different categories, which is computed as the average Euclidean distance from all exemplars in each category to category centroids of all other categories. Mathematically, given a system of $N$ words, $W=\{w_1,w_2,\cdots,w_N\}$, with $P$ linguistic exemplars $L=\{l_{11},\cdots,l_{1P},l_{21},\cdots,l_{2P},\cdots,l_{NP}\}$ and $Q$ visual exemplars $V=\{v_{11},\cdots,v_{1Q},v_{21},\cdots,v_{2Q},\cdots,v_{NQ}\}$ for each category, linguistic category centroids $LC=\{lc_1,\cdots,lc_N\}$ and visual category centroids $VC=\{vc_1,\cdots,vc_N\}$ are computed as 
$$lc_i=\frac{1}{P}\sum^P_{j=1} l_{ij},\ vc_i=\frac{1}{Q}\sum^Q_{j=1} v_{ij}$$
Linguistic variability and visual variability are then defined as 
$$lv_i=\frac{1}{NP} \sum^N_{i=1} \sum^P_{j=1} dist(l_{ij},lc_i),$$
$$vv_i=\frac{1}{NQ} \sum^N_{i=1} \sum^Q_{j=1} dist(v_{ij},vc_i)$$
while linguistic and visual discriminability are defined as 
$$ld_i=\frac{1}{N^2P} \sum^N_{i=1} \sum ^N_{j=1} \sum^P_{k=1} dist(l_{jk},lc_i),$$ 
$$vd_i=\frac{1}{N^2Q} \sum^N_{i=1} \sum ^N_{j=1} \sum^Q_{k=1} dist(v_{jk},vc_i)$$
where $dist$ computes the Euclidean distance between two points in an embedding space.

\subsection*{Alignment Strength}
To estimate the alignment between words' linguistic and visual representations, we used a metric termed Alignment Strength, which asks whether words’ linguistic similarities are correlated with their visual similarities. Computationally, we first constructed the similarity matrices of representations within the visual modality, $S_V$, and language modality, $S_L$, respectively by computing the pairwise cosine similarity of representations. The alignment strength was then computed as Spearman's rank correlation between the upper triangle of $S_V$ and $S_L$.

The relative alignment strength is defined as the percentage of misaligned mappings that have lower alignment strength than the true mapping system. For each true mapping, we randomly sampled $1,000$ mappings that were misaligned (e.g., the visual ``dog'' mapped to language ``cat'' while the visual ``cat'' mapped to language ``dog''), and then computed the alignment strength of those misaligned systems as a reference for the true mapping.

We randomly sampled $1,000$ systems to estimate the alignment strength of systems with 1 visual exemplar and 1 linguistic exemplar per category (Figure \ref{fig:alignment}A). Similarly, we randomly sampled $1,000$ systems to estimate the alignment strength of systems with 8 visual exemplars and 8 linguistic exemplars per category (Figure \ref{fig:alignment}D).

\subsection*{Exemplar Aggregation} Exemplar aggregation was achieved by taking the arithmetic mean of exemplars from the same category. For visual aggregation, we first created stable linguistic prototypes from 20 linguistic exemplars per category (post-hoc experiments confirmed that 20 exemplars per category can form well-learned linguistic prototypes). Instead of forming systems by purely random sampling, we incrementally added one visual exemplar to the existing system and compared the alignment strength of the true system with 1,000 permuted mappings to simulate the learning process of humans. We ran 1,000 simulations for each level of number of visual exemplars. The manipulation of linguistic aggregation was identical to visual aggregation, where the roles of visual modality and language modality were exchanged. 

The procedure for two-dimensional aggregation was also similar, where 500 simulations were run for each possible combination of  number of visual exemplars and number of linguistic exemplars.

\subsection*{Regression Analysis}
AoAs were measured directly from children's utterances in child-directed speech by vocabulary data retrieved from the Wordbank database \cite{frank2017wordbank}. Estimates of the frequency of words that children hear were based on parental/caregivers' speech transcripts from the CHILDES database \cite{macwhinney2014childes}.

Instead of conventional linear models, we chose a type of gradient-boosted trees model called XGBoost \cite{chen2016xgboost} for the regression analysis because tree-based models make fewer assumptions about the relationship between input and output and are immune to the issue of multi-collinearity, which is especially important when a large number of predictor variables are involved. The XGBoost model was trained for 10,000 runs with a max depth of 10 and a learning rate of 0.02.

\subsection*{Data, Materials, and Software Availability} Data and code for the current  study are available through the GitHub  repository: \href{https://github.com/FlamingoZh/verb_learning_quantification}{https://github.com/FlamingoZh/verb\_learning\_quantification}
}

\showmatmethods{} 

\acknow{This work was supported by the James S. McDonnell Foundation. We thank Leila Wehbe, Brian MacWhinney, Erik Thiessen, and Graham Newbig for their helpful comments.}

\showacknow{} 

\bibliography{ref}

\end{document}